\documentclass[conference]{IEEEtran}

\usepackage{times}

\usepackage{multicol}
\usepackage[colorlinks=true,urlcolor=red,pagebackref]{hyperref}

\usepackage{flushend}
\newcommand\inputpgf[2]{{
\let\pgfimageWithoutPath\pgfimage
\renewcommand{\pgfimage}[2][]{\pgfimageWithoutPath[##1]{#1/##2}}
\input{#1/#2}
}}

\usepackage{tabularx}
\newcolumntype{Y}{>{\centering\arraybackslash}X} %

\newcommand{\specialcell}[2][c]{%
  \begin{tabular}[#1]{@{}c@{}}#2\end{tabular}}
  
\usepackage{caption}

\usepackage{comment}
\usepackage{siunitx}
\usepackage{relsize}
\usepackage{ifthen}
\usepackage[colorinlistoftodos]{todonotes}

\usepackage[caption=false]{subfig}

\usepackage[vlined,ruled,linesnumbered]{algorithm2e}
\usepackage{graphics} %
\usepackage{rotating}
\usepackage{color}
\usepackage{enumerate}
\usepackage[T1]{fontenc}
\usepackage{psfrag}
\usepackage{epsfig} %
\usepackage{booktabs}
\usepackage{graphicx,url}
\usepackage{multirow}
\usepackage{array}
\usepackage{latexsym}
\usepackage{amsfonts}
\usepackage{amsmath}
\usepackage{amssymb}
\usepackage{xstring}
\usepackage[noend]{algorithmic}
\usepackage{multirow}
\usepackage{xcolor}
\usepackage{prettyref}
\usepackage{flexisym}
\usepackage{bigdelim}
\usepackage{breqn} %
\usepackage{listings}
\let\labelindent\relax
\usepackage{enumitem}
\usepackage{xspace}
\usepackage{bm}
\graphicspath{{./figures/}}
\usepackage{tikz}
\usetikzlibrary{matrix,calc}

\usepackage{mdwlist}

\let\itemize\undefined
\makecompactlist{itemize}{stditemize}

\newrefformat{prob}{Problem\,\ref{#1}}
\newrefformat{def}{Definition\,\ref{#1}}
\newrefformat{sec}{Section\,\ref{#1}}
\newrefformat{sub}{Section\,\ref{#1}}
\newrefformat{prop}{Proposition\,\ref{#1}}
\newrefformat{app}{Appendix\,\ref{#1}}
\newrefformat{alg}{Algorithm\,\ref{#1}}
\newrefformat{cor}{Corollary\,\ref{#1}}
\newrefformat{thm}{Theorem\,\ref{#1}}
\newrefformat{lem}{Lemma\,\ref{#1}}
\newrefformat{fig}{Fig.\,\ref{#1}}
\newrefformat{tab}{Table\,\ref{#1}}

\newtheorem{theorem}{Theorem}

\newtheorem{remark}[theorem]{Remark}

\newcommand{\bdmath}{\begin{dmath}}
\newcommand{\edmath}{\end{dmath}}
\newcommand{\beq}{\begin{equation}}
\newcommand{\eeq}{\end{equation}}
\newcommand{\bdm}{\begin{displaymath}}
\newcommand{\edm}{\end{displaymath}}
\newcommand{\bea}{\begin{eqnarray}}
\newcommand{\eea}{\end{eqnarray}}
\newcommand{\beal}{\beq \begin{array}{ll}}
\newcommand{\eeal}{\end{array} \eeq}
\newcommand{\beas}{\begin{eqnarray*}}
\newcommand{\eeas}{\end{eqnarray*}}
\newcommand{\ba}{\begin{array}}
\newcommand{\ea}{\end{array}}
\newcommand{\bit}{\begin{itemize}}
\newcommand{\eit}{\end{itemize}}
\newcommand{\ben}{\begin{enumerate}}
\newcommand{\een}{\end{enumerate}}

\newcommand{\setal}{~\emph{et~al.}\xspace}
\newcommand{\eg}{\emph{e.g.,}\xspace}
\newcommand{\ie}{\emph{i.e.,}\xspace}
\newcommand{\myParagraph}[1]{{\bf #1.}\xspace}

\newcommand{\hide}[1]{}

\newcommand{\hiddenText}{{\color{gray} hidden text.}}
\newcommand{\hideWithText}[1]{\hiddenText}

\newcommand{\at}[1]{^{(#1)}}

\newcommand{\scenario}[1]{{\smaller \sf#1}\xspace}

\newcommand{\blue}[1]{{\color{blue}#1}}

\newcommand{\linkToPdf}[1]{\href{#1}{\blue{(pdf)}}}
\newcommand{\linkToPpt}[1]{\href{#1}{\blue{(ppt)}}}
\newcommand{\linkToCode}[1]{\href{#1}{\blue{(code)}}}
\newcommand{\linkToWeb}[1]{\href{#1}{\blue{(web)}}}
\newcommand{\linkToVideo}[1]{\href{#1}{\blue{(video)}}}
\newcommand{\linkToMedia}[1]{\href{#1}{\blue{(media)}}}
\newcommand{\award}[1]{\xspace} %

\newcommand{\DSG}{\scenario{DSG}}
\newcommand{\DSGs}{\scenario{DSGs}}
\newcommand{\SPES}{\scenario{SPIN}}
\newcommand{\SPESs}{\scenario{SPINs}}
\newcommand{\TEASERpp}{\scenario{TEASER++}}
\newcommand{\SPESlong}{Spatial PerceptIon eNgine\xspace}
\newcommand{\SPESlongTwo}{spatial perception engine\xspace}

\newcommand{\Kimera}{{Kimera}\xspace}
\newcommand{\KimeraVIO}{{Kimera-VIO}\xspace}
\newcommand{\KimeraDVIOsuff}{DVIO}
\newcommand{\KimeraDVIO}{{\KimeraDVIOsuff}\xspace}
\newcommand{\KimeraRPGO}{{Kimera-RPGO}\xspace}
\newcommand{\KimeraMesher}{{Kimera-Mesher}\xspace}
\newcommand{\KimeraSemantics}{{Kimera-Semantics}\xspace}

\newcommand{\Euroc}{EuRoC\xspace}

\newcommand{\FigFrontCover}{Fig.~\ref{fig:frontCover}}%

\newcommand{\layerOneName}{Metric-Semantic Mesh\xspace}
\newcommand{\layerTwoName}{Objects and Agents\xspace}
\newcommand{\layerThreeName}{Places and Structures\xspace}
\newcommand{\layerFourName}{Rooms\xspace}
\newcommand{\layerSixName}{Building\xspace}

\newcommand{\maxJointDist}{\toCheck{3\rm{m}}}
\newcommand{\deltaT}{\toCheck{1 second}}

\newcommand{\UnityHumans}{\scenario{uHumans}}
\newcommand{\UnityOne}{\scenario{uH\_01}}
\newcommand{\UnityTwo}{\scenario{uH\_02\xspace}}
\newcommand{\UnityThree}{\scenario{uH\_03\xspace}}
\newcommand{\ESDF}{ESDF\xspace}
\newcommand{\toCheck}[1]{#1\xspace}

\newcommand{\optional}[1]{}%
\newcommand{\veryOptional}[1]{}%
\newcommand{\hideout}[1]{}

\usepackage{xcolor}
\usepackage{soul} %

\newcommand{\ourThe}{our~} %

\usepackage[numbers,sort&compress]{natbib}
\renewcommand{\baselinestretch}{0.97}%

\begin{document}

\def\Name{Kimera-X}
\title{3D Dynamic Scene Graphs: Actionable Spatial Perception with 
Places, Objects, and Humans}

\author{\authorblockN{Antoni Rosinol, Arjun Gupta, Marcus Abate, Jingnan Shi, Luca Carlone}
\authorblockA{
Laboratory for Information \& Decision Systems (LIDS) \\
Massachusetts Institute of Technology \\
\{arosinol,agupta,mabate,jnshi,lcarlone\}@mit.edu}
}

\makeatletter
\let\@oldmaketitle\@maketitle%
\renewcommand{\@maketitle}{
\@oldmaketitle%

  \begin{minipage}{\textwidth}
    \begin{center}
        \begin{tabular}{cc}
        \hspace{-2mm}\includegraphics[trim={0cm 0cm 0cm 0cm}, clip, width=0.99\columnwidth]{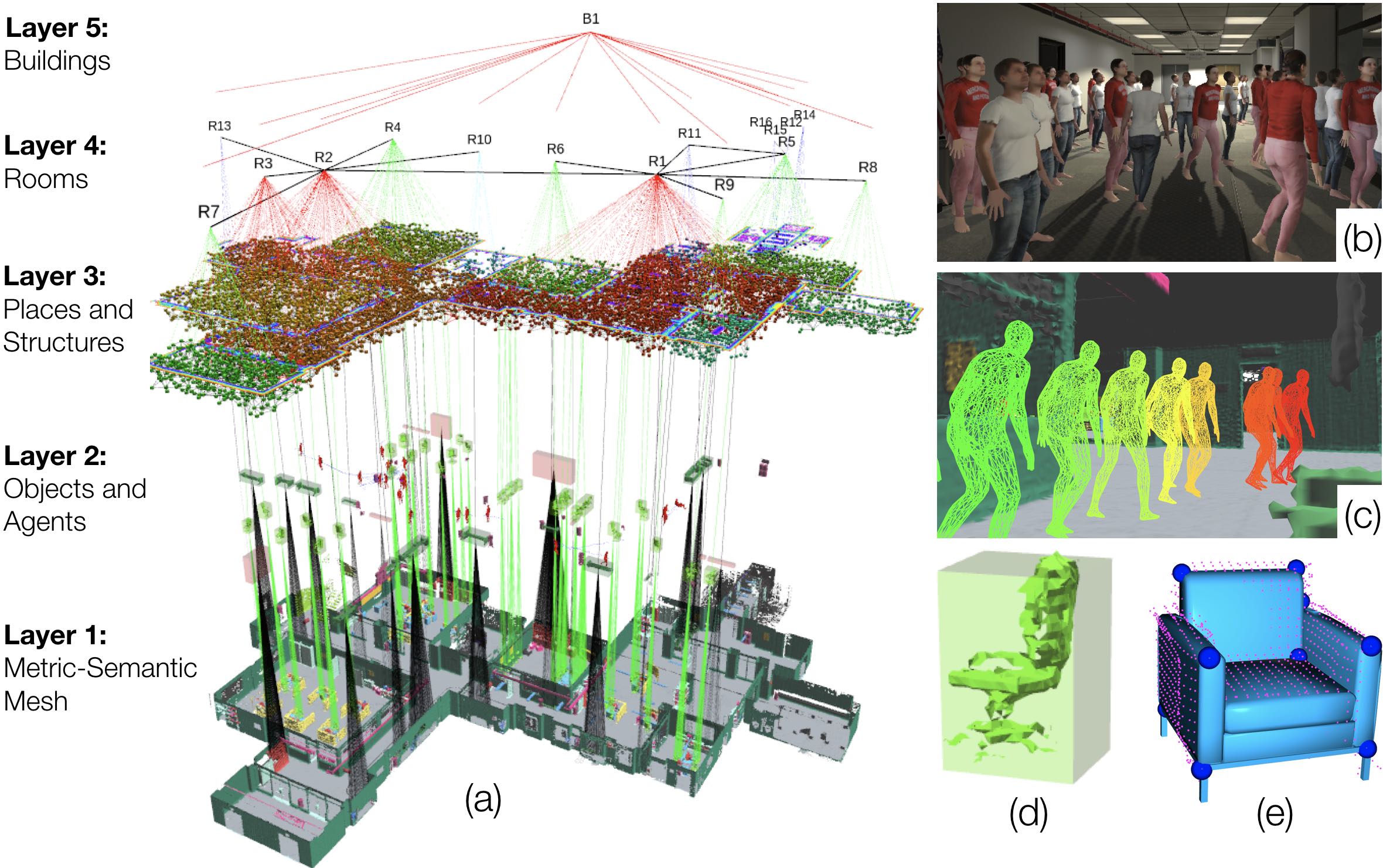}
        \end{tabular}
    \end{center}
    \vspace{-2mm}
  \end{minipage}
\\
\captionof{figure}{We propose {\bf 3D Dynamic Scene Graphs} (\DSGs) as a unified representation for actionable spatial perception. (a) A \DSG is a layered and hierarchical representation that abstracts a dense 3D model (\eg a metric-semantic mesh) into higher-level \emph{spatial concepts} (\eg objects, agents, places, rooms) 
and models their spatio-temporal relations (e.g., ``agent A is in room B at time $t$'', traversability between places or rooms). %
We present a \SPESlong (\SPES) that reconstructs a \DSG from visual-inertial data, 
and  
(a) segments places, structures (\eg walls), and rooms,
(b) is robust to extremely crowded environments,
(c) tracks dense mesh models of human agents in real time, 
(d) estimates centroids and bounding boxes of objects of unknown shape, 
(e) estimates the 3D pose of objects for which a CAD model is given. 
 \vspace{-4mm}}
 \label{fig:frontCover}
}%
\makeatother

\maketitle

\setcounter{figure}{1}

\begin{abstract}
We present a unified 
representation for actionable spatial perception:
\emph{3D Dynamic Scene Graphs}. 
\emph{Scene graphs} are directed graphs where nodes represent entities in the scene
 (\eg objects, walls, rooms), and edges represent relations (\eg inclusion, adjacency) among nodes.
{\emph{Dynamic} scene graphs (\DSGs) extend this notion  
to represent dynamic scenes with moving \emph{agents} (\eg humans, robots), 
and to include actionable information that supports planning and decision-making 
(\eg spatio-temporal relations, topology at different levels of abstraction).}
Our second contribution is to provide the first fully automatic \emph{\SPESlong} (\SPES) to build a \DSG from visual-inertial data. %
We integrate state-of-the-art techniques for object and human detection and pose estimation, 
and we describe how to robustly infer object, robot, and human nodes in crowded scenes. 
 To the best of our knowledge, 
this is the first paper that reconciles visual-inertial SLAM and dense human mesh tracking.
Moreover, we provide algorithms to obtain hierarchical representations of indoor environments (\eg places, structures, rooms) and their relations. %
Our third contribution is to demonstrate the proposed \SPESlongTwo in a 
photo-realistic Unity-based simulator, where we assess its robustness and expressiveness. 
Finally, we discuss the implications of our proposal on 
modern robotics applications. 
{3D Dynamic Scene Graphs can  have a profound impact on 
planning and decision-making, 
human-robot interaction, 
 long-term autonomy, and scene prediction.
 A video abstract is available at {\small\href{https://youtu.be/SWbofjhyPzI}{https://youtu.be/SWbofjhyPzI}}.
 }
\end{abstract} 
\IEEEpeerreviewmaketitle

\section{Introduction}
\label{sec:intro}

Spatial perception and 3D environment understanding are key enablers for high-level task execution in the real world. 
In order to execute high-level instructions, such as ``search for survivors on the second floor of the tall building'', 
a robot needs to 
ground semantic concepts (survivor, floor, building) into a spatial representation (\ie a metric map), 
leading to \emph{metric-semantic} spatial representations that go beyond the map models typically 
built by SLAM and visual-inertial odometry (VIO) pipelines~\cite{Cadena16tro-SLAMsurvey}. 
In addition, bridging low-level obstacle avoidance  and motion planning with high-level task planning requires 
 constructing a world model that captures reality at different levels of abstraction. For instance, while
 task planning might be effective in describing a sequence of actions to complete a task (\eg reach the entrance of the building, 
 take the stairs, enter each room), motion planning typically relies %
 on a fine-grained map representation (\eg a mesh or a volumetric model). 
 Ideally, spatial perception should be able to build a \emph{hierarchy of consistent abstractions} to feed both motion and task planning.
 {The problem becomes even more challenging  
when autonomous systems are deployed in crowded environments. From self-driving cars to collaborative robots on factory floors,  
identifying obstacles is not sufficient for safe and effective navigation/action, and 
 it becomes 
 crucial to reason on the \emph{dynamic} entities in the scene (in particular, \emph{humans}) 
 and predict their behavior or intentions~\cite{Everett18iros-motionAmongDynamicAgents}.}

The existing literature falls short of simultaneously addressing these issues (metric-semantic understanding, actionable hierarchical abstractions, modeling of dynamic entities). 
Early work on map representation in robotics (\eg~\cite{Kuipers00ai,Kuipers78cs,Chatila85,Vasudevan06iros,Galindo05iros-multiHierarchicalMaps,Zender08ras-spatialRepresentations}) 
 investigates hierarchical representations but mostly in 2D and assuming static environments; moreover, 
 these works were proposed before the ``deep learning revolution'', hence they could not afford advanced semantic 
 understanding. 
 On the other hand,
 the quickly growing literature on metric-semantic mapping (\eg~\cite{Salas-Moreno13cvpr,Bowman17icra,Behley19iccv-semanticKitti,Tateno15iros-metricSemantic,Rosinol19arxiv-Kimera,Grinvald19ral-voxbloxpp,McCormac17icra-semanticFusion}) 
 mostly focuses on ``flat'' representations (object constellations, metric-semantic meshes or volumetric models) that are not
  hierarchical in nature.  
  Very recent work~\cite{Armeni19iccv-3DsceneGraphs,Kim19tc-3DsceneGraphs} attempts to bridge this gap by designing richer representations, called \emph{3D Scene Graphs}.
 A scene graph is a data structure commonly used in computer graphics and gaming applications that consists of a 
 graph model where nodes represent entities in the scene and edges represent spatial or logical relationships among nodes.
 While the works~\cite{Armeni19iccv-3DsceneGraphs,Kim19tc-3DsceneGraphs} pioneered 
 the use of 3D scene graphs in robotics and vision (prior work in vision focused on 2D scene graphs defined in the image space~\cite{Choi13cvpr-sceneParsing,Zhao13cvpr-sceneParsing,Huang18eccv-sceneParsing, Jiang18ijcv-sceneParsing}), 
 they have important drawbacks. Kim\setal~\cite{Kim19tc-3DsceneGraphs} only capture objects and miss multiple levels of abstraction. Armeni\setal~\cite{Armeni19iccv-3DsceneGraphs} provide a hierarchical model that is useful for visualization and knowledge organization, but does not capture \emph{actionable} information, such as traversability, which is key to robot navigation. 
 Finally, %
 neither~\cite{Kim19tc-3DsceneGraphs} nor~\cite{Armeni19iccv-3DsceneGraphs}  account for or model dynamic entities in the environment.  

\myParagraph{Contributions} 
 We present a unified representation for actionable spatial perception:
\emph{3D Dynamic Scene Graphs} (\DSGs,~\FigFrontCover).
  A \DSG, introduced in Section~\ref{sec:DSG}, is a \emph{layered} directed graph where nodes represent
  \emph{spatial concepts} (\eg objects, rooms, agents) and edges represent pairwise spatio-temporal relations. 
   The graph is \emph{layered}, in that nodes are grouped into layers that correspond to different levels of abstraction of the scene 
   (\ie~a \DSG is a hierarchical representation).
   Our choice of nodes and edges in the \DSG also captures \emph{places} and their connectivity, hence providing a 
   strict generalization of the notion of topological maps~\cite{Ranganathan04iros,Remolina04} and making \DSGs an \emph{actionable} representation for navigation and planning. 
   Finally, edges in the \DSG capture spatio-temporal relations and explicitly model dynamic entities in the scene, and in particular humans, for which we estimate both 3D poses over time (using a \emph{pose graph} model) and a mesh model. 

Our second contribution, presented in Section~\ref{sec:buildingDSG}, 
is to provide the first fully automatic \emph{\SPESlong} (\SPES) 
to build a \DSG. %
 While the state of the art~\cite{Armeni19iccv-3DsceneGraphs} assumes an annotated mesh model of the environment is given and relies on a semi-automatic procedure to extract the scene graph, we present a pipeline that starts from 
 visual-inertial data and builds the \DSG without human supervision. 
 Towards this goal (i) we integrate state-of-the-art techniques for object~\cite{Yang20arxiv-teaser} 
 and human~\cite{Kolotouros19cvpr-shapeRec} detection and pose estimation, 
 (ii) we describe how to robustly infer object, robot, and human nodes in cluttered and crowded scenes,  
 and (iii) we provide algorithms to partition an indoor environment into places, structures, and rooms. 
This is the first paper that integrates visual-inertial SLAM and human mesh tracking 
(we use SMPL meshes~\cite{Loper15tg-smpl}).
\emph{The notion of 
\SPES generalizes SLAM, which becomes a module in our pipeline, and augments it to capture  
relations, dynamics, and high-level abstractions.}

Our third contribution, in Section~\ref{sec:experiments}, is to demonstrate the proposed \SPESlongTwo  in a 
 Unity-based photo-realistic simulator, where we assess its robustness and expressiveness. 
We show that our \SPES 
(i) includes desirable features that improve the robustness of mesh reconstruction and
human tracking (drawing connections with the literature on pose graph optimization~\cite{Cadena16tro-SLAMsurvey}),
 (ii) can deal with both objects of known and unknown shape, and 
 (iii) uses a simple-yet-effective  
 heuristic to segment places and rooms in an indoor environment. 
 More extensive and interactive visualizations are given in the video attachment (available at \cite{Rosinol20website-uHumans}).

Our final contribution, in Section~\ref{sec:discussion}, is to discuss several \emph{queries} a \DSG can support, and its use as an \emph{actionable} 
spatial perception model. In particular, 
we discuss how \DSGs can impact  
planning and decision-making (by providing a representation for hierarchical planning and fast collision checking), 
human-robot interaction (by providing an interpretable abstraction of the scene), 
 long-term autonomy (by enabling data compression), and scene prediction.
\veryOptional{We believe future research on \DSGs and \SPESs can bridge the gap between robotics, computer vision, and graphics, 
enabling to build a rich and actionable 3D model in real-time, that can be then used within a physics simulator to 
roll out potential evolutions of the scene (see also~\cite{Zheng13cvpr-sceneUnderstanding}), and support human-level scene understanding and decision-making.}

\section{Related Work}
\label{sec:relatedWork}

\myParagraph{Scene Graphs}
Scene graphs are popular computer graphics models to describe, manipulate, and render complex scenes and are commonly used in game engines~\cite{Wang10book-openSceneGraph}.
 While in gaming applications, these structures are used to describe 3D environments,
 scene graphs have been mostly used in computer vision to abstract the content of 2D images.
Krishna\setal~\cite{Krishna16arxiv-visualGenome} use a scene graph to model attributes and relations among objects in 2D images, relying on manually defined natural language captions.
 Xu\setal~\cite{Xu17iccv-sceneGraphGeneration} and Li\setal~\cite{Li17iccv-sceneGraphGeneration} develop algorithms for 2D scene graph generation.
 2D scene graphs have been used for image retrieval~\cite{Johnson15cvpr-sceneGraphImageRetrieval},
 captioning~\cite{Krause17cvpr-sceneGraphDescription,Anderson16eccv-sceneGraphDescription,Johnson17cvpr-sceneGraphDescription},
 high-level understanding~\cite{Choi13cvpr-sceneParsing,Zhao13cvpr-sceneParsing,Huang18eccv-sceneParsing, Jiang18ijcv-sceneParsing}, %
 visual question-answering~\cite{Fukui16arxiv-sceneGraphQandA,Zhu16cvpr-sceneGraphQandA},
and action detection~\cite{Lu16cvpr-sceneGraphRelations,Liang17cvpr-sceneGraphRelations,Zhang17cvpr-sceneGraphRelations}.

Armeni\setal~\cite{Armeni19iccv-3DsceneGraphs} propose a \emph{3D scene graph} model to describe 3D static scenes,
 and describe a semi-automatic algorithm to build the scene graph.
 In parallel to~\cite{Armeni19iccv-3DsceneGraphs}, Kim\setal~\cite{Kim19tc-3DsceneGraphs} propose a 3D scene graph model for robotics, which however only includes objects as nodes and misses multiple levels of abstraction afforded by~\cite{Armeni19iccv-3DsceneGraphs} and by our proposal.
 \veryOptional{Contrarily to~\cite{Armeni19iccv-3DsceneGraphs,Kim19tc-3DsceneGraphs}, which propose a static representation of the environment,  we propose a dynamic and actionable representation, and automatically infer it from visual-inertial data.}

\myParagraph{Representations and Abstractions in Robotics}
 The question of world modeling and map representations has been central in the robotics community since its
inception~\cite{Thrun02a,Cadena16tro-SLAMsurvey}.
The need to use hierarchical maps that capture rich spatial and semantic information was already
recognized in seminal papers by Kuipers, Chatila, and Laumond~\cite{Kuipers00ai,Kuipers78cs,Chatila85}.
Vasudevan\setal~\cite{Vasudevan06iros} propose a hierarchical representation of object constellations.
 Galindo\setal~\cite{Galindo05iros-multiHierarchicalMaps} use two parallel hierarchical representations
(a spatial and a semantic representation) that are then \emph{anchored} to each other and estimated using 2D lidar data.
 Ruiz-Sarmiento\setal~\cite{Ruiz-Sarmiento17kbs-multiversalMaps} extend the framework in~\cite{Galindo05iros-multiHierarchicalMaps} to account for uncertain groundings between spatial and semantic elements.
  Zender\setal~\cite{Zender08ras-spatialRepresentations} propose a single hierarchical representation that includes a 2D map, a navigation graph and a topological map~\cite{Ranganathan04iros,Remolina04}, which are then further abstracted into a \emph{conceptual map}.
Note that the spatial hierarchies in~\cite{Galindo05iros-multiHierarchicalMaps} and~\cite{Zender08ras-spatialRepresentations} already resemble a scene graph, with
less articulated set of nodes and layers. A more fundamental difference is the fact that
early work
(i) did not reason over 3D models (but focused on 2D occupancy maps),
(ii) did not tackle dynamical scenes, and
(iii) did not include dense (e.g., pixel-wise) semantic information, which
has been enabled in recent years by deep learning methods.
\myParagraph{Metric-Semantic Scene Reconstruction}
This line of work is concerned with estimating metric-semantic (but typically non-hierarchical) representations from sensor data.
While early work~\cite{Bao11cvpr,Brostow08eccv} focused on offline processing,
recent years have seen a surge of interest towards \emph{real-time} metric-semantic mapping, %
 triggered by pioneering works such as SLAM++~\cite{Salas-Moreno13cvpr}.
\emph{Object-based approaches} compute an object map and include SLAM++~\cite{Salas-Moreno13cvpr},
 XIVO~\cite{Dong17cvpr-XVIO}, OrcVIO~\cite{Mo19tr-orcVIO}, QuadricSLAM~\cite{Nicholson18ral-quadricSLAM},
 and~\cite{Bowman17icra}.
 For most robotics applications, an object-based map
 does not provide enough resolution for navigation and obstacle avoidance.
\emph{Dense approaches} build denser semantically annotated models in the form
of point clouds~\cite{Behley19iccv-semanticKitti,Tateno15iros-metricSemantic,Renaud18rss-segMap,Lianos18eccv-VSO},
meshes~\cite{Rosinol19arxiv-Kimera,Grinvald19ral-voxbloxpp,Rosu19ijcv-semanticMesh},  %
surfels~\cite{Whelan15rss-elasticfusion,Wald18ral-metricSemantic},
or volumetric models~\cite{McCormac17icra-semanticFusion,Grinvald19ral-voxbloxpp,Narita19arxiv-metricSemantic}.
Other approaches use both objects and dense models, see
Li\setal~\cite{Li16iros-metricSemantic} and
  Fusion++~\cite{McCormac183dv-fusion++}.
 These approaches focus on static environments.
  Approaches that deal with moving objects, such as DynamicFusion~\cite{Newcombe15cvpr-dynamicFusion},
  Mask-fusion~\cite{Runz18ismar-maskfusion},
  Co-fusion~\cite{Runz17icra-cofusion}, and MID-Fusion~\cite{Xu19icra-midFusion}
 are currently limited to small table-top scenes and focus on objects or dense maps, rather than  scene graphs.
\myParagraph{Metric-to-Topological Scene Parsing}
This line of work focuses on %
partitioning a metric map into semantically meaningful places (\eg rooms, hallways).
N{\"u}chter and Hertzberg~\cite{Nuchter08ras-semanticMaps} encode relations among planar surfaces (\eg walls, floor, ceiling) and detect objects in the scene.
Blanco\setal~\cite{Blanco09ras-metricTopologicalSLAM} propose a hybrid metric-topological map.
Friedman\setal~\cite{Friedman07ijcai-voronoiRF} propose \emph{Voronoi Random Fields} to obtain an abstract model of a 2D
grid map.
 Rogers and Christensen~\cite{Rogers12icra-semanticMapping} and Lin\setal~\cite{Lin13cvpr-holisticSceneUnderstanding}
 leverage objects to perform a joint object-and-place classification.
Pangercic\setal~\cite{Pangercic12icra-objectMaps} reason on the objects' functionality.
Pronobis and Jensfelt~\cite{Pronobis12icra-MRFsemanticMapping} use a Markov Random Field to segment a 2D grid map.
Zheng\setal~\cite{Zheng18aaai-graphStructuredNet}
infer the topology of a grid map using a \emph{Graph-Structured Sum-Product Network}, while Zheng and Pronobis~\cite{Zheng19iros-topoNet} use a neural network.
Armeni\setal~\cite{Armeni16cvpr-3DsemanticParsing} focus on a 3D mesh, and propose a method to parse a building into rooms. %
Floor plan estimation has been also investigated using single images~\cite{Hedau09cvpr-floorPlan},
omnidirectional images~\cite{Lukierski17icra-floorPlan}, 2D lidar~\cite{Li20arxiv-floorPlan,Turner14grapp-floorPlan},
3D lidar~\cite{Mura14cg-floorPlan,Ochmann14grapp-floorPlan}, RGB-D~\cite{Chen18eccv-floorNet},
or from crowd-sourced mobile-phone trajectories~\cite{Alzantot12icagis-floorPlan}.
The works~\cite{Armeni16cvpr-3DsemanticParsing,Mura14cg-floorPlan,Ochmann14grapp-floorPlan} are closest to our proposal,
but contrarily to~\cite{Armeni16cvpr-3DsemanticParsing} we do not rely on a Manhattan World assumption,
and contrarily to~\cite{Mura14cg-floorPlan,Ochmann14grapp-floorPlan} we operate on a mesh model.
\myParagraph{SLAM and VIO in Dynamic Environments}
This paper is also concerned with modeling and gaining robustness against dynamic elements in the scene.
SLAM and moving object tracking
has been extensively investigated in robotics~\cite{Wang07ijrr-slammot, Azim12ivs-datmo},
 while more recent work focuses on joint visual-inertial odometry and target pose
 estimation~\cite{Qiu19tro-vioObjectTracking,Eckenhoff19ral-vioObjectTracking,Geneva19arxiv-VIOandTargetTracking}.
Most of the existing literature in robotics models the
dynamic targets as a single 3D point \cite{Chojnacki18ijmav-motionAndObjectTracking},
or with a 3D pose and rely on lidar~\cite{Azim12ivs-datmo}, RGB-D cameras~\cite{Aldoma13icra-objectTracking},
monocular cameras~\cite{Peiliang18eccv-motionAndObjectTracking},
and visual-inertial sensing~\cite{Qiu19tro-vioObjectTracking}.
  Related work also attempts to gain robustness against dynamic scenes by using IMU motion information~\cite{Hwangbo09iros-IMUKLT},
  or masking portions of the scene corresponding to dynamic elements~\cite{Cui19access-SOFSLAM,Brasch18iros-dynamicSLAM,Bescos18ral-dynaSLAM}.
To the best of our knowledge, the present paper is the first work that attempts to perform visual-inertial SLAM,
 segment dense object models, estimate the 3D poses of known objects, and
 reconstruct and track dense human SMPL meshes.
\myParagraph{Human Pose Estimation}
Human pose and shape estimation from a single image is a growing research area. While
we refer the reader to~\cite{Kolotouros19cvpr-shapeRec,Koloturos19arxiv-IterativeShape} %
for a broader review, it is worth mentioning that related work includes optimization-based approaches,
 which fit a 3D mesh to 2D image keypoints~\cite{Bogo16eccv-keepItSMPL,Lassner17cvpr-humanPoseAndShape,Zanfir18cvpr-humanPoseAndShape,Koloturos19arxiv-IterativeShape,Yang19arxiv-shapeStar}, and learning-based methods,
 which infer the mesh directly from pixel information~\cite{Tan17bmvc-humanPoseAndShape,Kanazawa18cvpr-humanPoseAndShape,Omran183dv-humanPoseAndShape,Pavlakos18cvpr-humanPoseAndShape,Kolotouros19cvpr-shapeRec,Koloturos19arxiv-IterativeShape}.
 Human models are typically parametrized using the \emph{Skinned Multi-Person Linear Model} (SMPL)~\cite{Loper15tg-smpl}, which provides a compact
  pose and shape description and can be rendered as a mesh with 6890 vertices and 23 joints.

\section{3D Dynamic Scene Graphs}
\label{sec:DSG}

A 3D \emph{Dynamic Scene Graph} (\DSG, \FigFrontCover) is an actionable spatial representation that captures the 3D geometry and semantics of a scene at different levels of abstraction,
 and models objects, places, structures, and agents and their relations. 
 More formally, a \DSG is a \emph{layered directed graph} where nodes represent
  \emph{spatial concepts} (\eg objects, rooms, agents) and edges represent pairwise spatio-temporal relations (\eg ``agent A is in room B at time $t$''). 
  Contrarily to \emph{knowledge bases}~\cite{Krishna92book-knowledgeBase}, spatial concepts are semantic concepts that are \emph{spatially grounded} (in other words, each node in our \DSG includes spatial coordinates and shape or bounding-box information as attributes). 
  A \DSG is a \emph{layered} graph, \ie nodes are grouped into layers that correspond to different levels of abstraction.  Every node has a unique ID.

The \DSG of a single-story indoor environment includes 5 layers (from low to high abstraction level): 
(i) \layerOneName, 
(ii) \layerTwoName,
(iii) \layerThreeName, 
(iv) \layerFourName, 
and 
(v) \layerSixName.
We discuss each layer and the corresponding nodes and edges below.
\subsection{Layer 1: \layerOneName} %
\label{sec:layer-mesh}

The lower layer of a \DSG is a semantically annotated 3D mesh (bottom of \FigFrontCover(a)). 
The nodes in this layer are 3D points (vertices of the mesh) and each node has the following attributes:
(i) 3D position, (ii) normal, (iii) RGB color, and (iv) a panoptic semantic label.\footnote{Panoptic segmentation~\cite{Kirillov19cvpr-panopticSegmentation,Li18arxiv-thingsAndStuff} segments both object (\eg chairs, tables, drawers) instances and structures (\eg walls, ground, ceiling).} 
Edges connecting triplets of points (\ie a clique with 3 nodes) describe faces in the mesh and define the 
topology of the environment. %
Our metric-semantic mesh includes everything in the environment that is \emph{static}, while for storage convenience we 
store meshes of dynamic objects in a separate structure (see ``Agents'' below).

\newcommand{\myhspace}{\hspace{-6mm}}
\newcommand{\mpw}{4.5cm}
\newcommand{\myRate}{1}

\begin{figure}[t]
\vspace{-3mm}
	\begin{center}
	\begin{minipage}{\columnwidth}
	\includegraphics[trim=0mm 1mm 0mm 0mm, clip, width=\myRate\columnwidth]{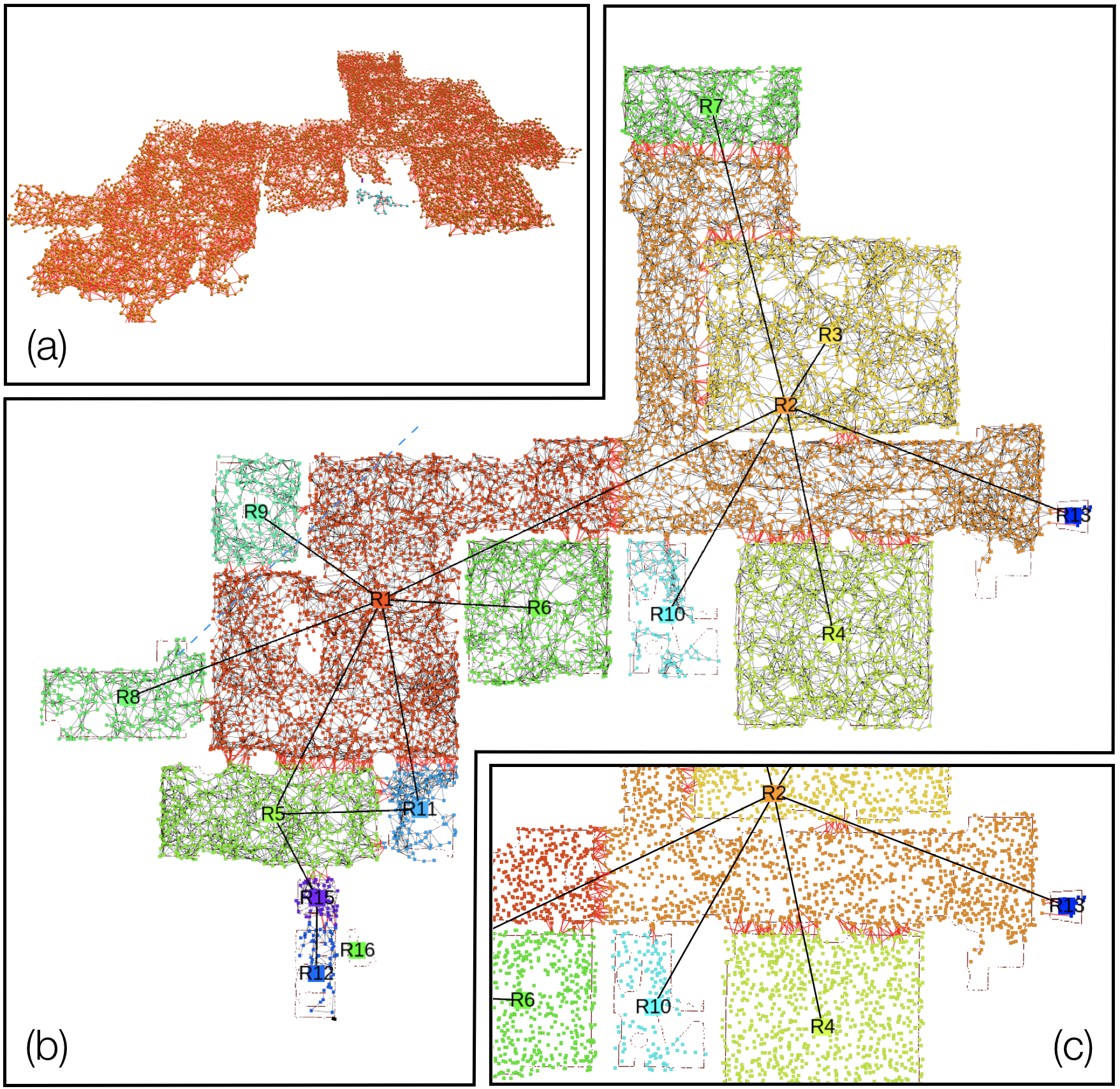} \\
	\end{minipage}
	\vspace{-3mm} 
	\caption{
	{\bf Places} and their connectivity shown as a graph.
	(a) Skeleton (places and topology) produced by~\cite{Oleynikova18iros-topoMap} (side view);
	 (b) {\bf Room} parsing produced by our approach 
	 (top-down view);
	 (c) Zoomed-in view; red edges connect different rooms.
	 \label{fig:rooms}\vspace{-4mm} }
	\end{center}
	\vspace{-2mm}
\end{figure} 
\subsection{Layer 2: \layerTwoName}
\label{sec:layer-objectsAgents}

This layer contains two types of nodes: objects and agents (\FigFrontCover(c-e)), whose main distinction is the fact that agents are time-varying entities, while 
objects are static.

{\bf Objects} represent static elements in the environment that are not considered \emph{structural} (\ie walls, floor, ceiling, pillars are considered \emph{structure} and are not modeled in this layer). 
Each object is a node and node attributes include (i) a 3D object pose,
\veryOptional{\footnote{3D poses of static objects are sometimes called ``Static Coordinate Systems'' in  graphics~\cite{Rohlf94accgit-iris}.}} (ii) a bounding box, and (ii) its semantic class (\eg chair, desk).
While not investigated in this paper, we refer the reader to~\cite{Armeni19iccv-3DsceneGraphs} for a more comprehensive list of attributes, including materials and affordances.
Edges between objects describe relations, such as  co-visibility, relative size, distance, or contact (``the cup is on the desk'').
Each object node is connected to the corresponding set of points belonging to the object in the \layerOneName. 
Moreover, nearby objects are connected to the same \emph{place} node (see Section~\ref{sec:layer-places}). 

{\bf Agents} represent dynamic entities in the environment, including humans. 
While in general there might be many types of dynamic entities (\eg vehicles, bicycles in outdoor environments), without loss of generality 
here we focus on two classes: \emph{humans} and \emph{robots}.\footnote{These classes can be considered instantiations of more general concepts: 
``rigid'' agents (such as robots, for which we only need to keep track a 3D pose), and ``deformable'' agents 
(such as humans, for which we also need to keep track of a 
time-varying shape).}
 Both human and robot nodes have three attributes: 
 (i) a 3D pose graph describing their trajectory over time,
 \veryOptional{\footnote{3D poses of dynamic objects are sometimes called ``Dynamic Coordinate Systems'' in  graphics~\cite{Rohlf94accgit-iris}.}} 
 (ii) a mesh model describing their (non-rigid) shape, and
 (iii) a semantic class (\ie human, robot). 
 A pose graph~\cite{Cadena16tro-SLAMsurvey} is a collection of time-stamped 3D poses where edges 
 model pairwise relative measurements.
The robot collecting the data is also modeled as an agent in this layer.

\subsection{Layer 3: \layerThreeName}
\label{sec:layer-places}

This layer contains two types of nodes: places and structures. 
Intuitively, places are a model for the free space, while structures capture separators between different spaces.

{\bf Places} (Fig.~\ref{fig:rooms}) correspond to positions in the free-space and edges between places represent traversability (in particular: presence of a straight-line path between places).
Places and their connectivity form a \emph{topological map}~\cite{Ranganathan04iros,Remolina04} that can be used for path planning.
\optional{Moreover, place nodes provide more granularity in the scene description (\eg the \DSG can differentiate the back of the room from its front).} %
Place attributes only include a 3D position, but can also include a semantic class (\eg back or front of the room) and an obstacle-free bounding box around the place position.
Each object and agent in Layer 2 is connected with the nearest place (for agents, the connection is for each time-stamped pose, since 
agents move from place to place).
Places belonging to the same room are also connected to the same room node in Layer~4. 
Fig.~\ref{fig:rooms}(b-c) shows a visualization with places color-coded by rooms. %

{\bf Structures} (Fig.~\ref{fig:structure}) include nodes describing structural elements in the environment, \eg walls, floor, ceiling, pillars. 
The notion of structure captures elements often called  ``stuff'' in related work~\cite{Li18arxiv-thingsAndStuff}, while
 we believe the 
name ``structure'' is more evocative and useful to contrast them to objects. %
Structure nodes' attributes are: (i) 3D pose, (ii) bounding box, and (iii) semantic class (\eg walls, floor).
Structures may have edges to the rooms they enclose. 
Structures may also have edges to an object in Layer 3, \eg a ``frame'' (object) ``is hung'' (relation) on 
a ``wall'' (structure), or a ``ceiling light is mounted on the ceiling''. 

\renewcommand{\myhspace}{\hspace{-6mm}}
\renewcommand{\mpw}{4.5cm}
\renewcommand{\myRate}{1}

\begin{figure}[h]
	\begin{center}
	\begin{minipage}{\columnwidth}
	\includegraphics[trim=0mm 5mm 0mm 5mm, clip,width=0.9\myRate\columnwidth]{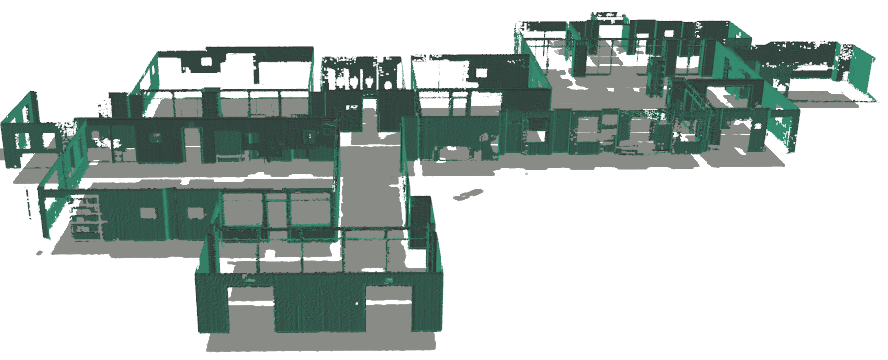} \\
	\end{minipage}
	\vspace{-5mm} 
	\caption{{\bf Structures}: exploded view of walls and floor. %
	 \label{fig:structure}}
	\vspace{-8mm} 
	\end{center}
\end{figure} 
\subsection{Layer 4: \layerFourName} %
\label{sec:layer-rooms}

This layer includes nodes describing rooms, corridors, and halls. 
Room nodes (Fig.~\ref{fig:rooms}) have the following attributes: (i) 3D pose, (ii) bounding box, and (iii) semantic class (\eg kitchen, dining room, corridor).
Two rooms are connected by an edge if they are adjacent (\ie there is a door connecting them).
A room node has edges to the %
places (Layer 3) it contains (since each place is connected to nearby objects, the \DSG also captures which object/agent is contained in each room).  
All rooms are connected to the building they belong to (Layer 5). 

\subsection{Layer 5: \layerSixName} %
\label{sec:layer-building}

Since we are considering a representation over a single building, there is a single \emph{building node} 
with the following attributes: (i) 3D pose, (ii) bounding box, and (iii) semantic class (\eg office building, 
residential house). The building node has edges towards all rooms in the building.

\subsection{Composition and Queries} %

\emph{Why should we choose this set of nodes or edges rather than a different one?}
 Clearly, the choice of nodes in the \DSG is not unique and is task-dependent.
  Here we first motivate our choice of nodes  in terms of \emph{planning queries} the \DSG is designed for
  (see Remark~\ref{rmk:queries} and the broader discussion in Section~\ref{sec:discussion}), and we then show that the representation is compositional, in the sense that 
  it can be easily expanded to encompass more layers, nodes, and edges (Remark~\ref{rmk:composition}).

\begin{remark}[Planning Queries]
\label{rmk:queries}

The proposed \DSG is designed with  task and motion planning queries in mind. 
The semantic node attributes (\eg semantic class) support planning from high-level specification 
(``pick up the red cup from the table in the dining room''). 
The geometric node attributes (\eg meshes, positions, bounding boxes) 
and the edges are used for motion planning. For instance, 
the places can be used as a topological graph for path planning, and the bounding boxes can be used for fast 
collision checking. \optional{A more extensive discussion about \DSG queries is postponed to Section~\ref{sec:discussion}.} 
\end{remark}

\begin{remark}[Composition of \DSGs]
\label{rmk:composition}

A second re-ensuring property of a \DSG is its compositionality: 
  one can easily concatenate more layers at the top and the bottom of the \DSG in \FigFrontCover(a), and 
  even add intermediate layers. 
  For instance, in a multi-story building, we can include a ``Level'' layer between the ``Building'' and ``Rooms'' layers in 
  \FigFrontCover(a). Moreover, we can add further abstractions or layers at the top, for instance going from 
  buildings to neighborhoods, and then to cities. 
\end{remark}

\section{Spatial Perception Engine: \\ Building a 3D \DSGs from Sensor Data}
\label{sec:buildingDSG}

This section describes a \emph{\SPESlong} (\SPES) that populates the \DSG nodes and edges using sensor data.
 The input to our \SPES is streaming data from a stereo camera and an Inertial Measurement Unit (IMU).
The output is a 3D \DSG. 
In our current implementation, the metric-semantic mesh and the agent nodes are incrementally built from sensor data in real-time,
while the remaining nodes (objects, places, structure, rooms) are automatically built at the end of the run.

 Section~\ref{sec:fromVItoMeshAndAgents} describes how to obtain the metric-semantic mesh and agent nodes 
 from sensor data. Section~\ref{sec:fromMeshToObjects} 
 describes how to segment and localize objects. Section~\ref{sec:fromMeshToStructures} describes how to 
 parse places, structures, and rooms.

\subsection{From Visual-Inertial data to Mesh and Agents}
\label{sec:fromVItoMeshAndAgents}

\myParagraph{Metric-Semantic Mesh}
We use \Kimera~\cite{Rosinol19arxiv-Kimera} to reconstruct a semantically annotated 3D mesh from visual-inertial data in real-time.
\Kimera %
is open source and includes four main modules: 
(i) \KimeraVIO: a visual-inertial odometry module implementing IMU preintegration and fixed-lag smoothing~\cite{Forster17tro}, 
(ii) \KimeraRPGO: a robust pose graph optimizer~\cite{Mangelson18icra}, 
(iii) \KimeraMesher: a per-frame and multi-frame mesher~\cite{Rosinol19icra-mesh}, 
and 
(iv) \KimeraSemantics: a volumetric approach to produce a semantically annotated mesh and an Euclidean Signed Distance Function (ESDF)
 based on Voxblox~\cite{Oleynikova2017iros-voxblox}.
 \KimeraSemantics uses a panoptic 2D semantic segmentation of the left camera images to label the 3D mesh using Bayesian updates. 
We take the metric-semantic mesh produced by \KimeraSemantics as Layer 1 in the \DSG in \FigFrontCover(a).

\myParagraph{Robot Node}
In our setup the only robotic agent is the one collecting the data, hence  \KimeraRPGO directly produces a time-stamped 
pose graph describing the poses of the robot at discrete time stamps. 
Since our robot moves in crowded environments, we replace the Lukas-Kanade tracker in the 
VIO front-end of~\cite{Rosinol19arxiv-Kimera} with an IMU-aware optical flow method, where feature motion between frames is predicted using 
IMU motion information, similar to~\cite{Hwangbo09iros-IMUKLT}. Moreover, we use a 2-point RANSAC~\cite{Kneip11bmvc} for geometric verification, which directly uses the IMU rotation to prune outlier correspondences in the feature tracks. 
To complete the robot node, we assume a CAD model of the robot to be given (only used for visualization).

\myParagraph{Human Nodes}
Contrary to related work that models dynamic targets as a point or a 3D pose~\cite{Chojnacki18ijmav-motionAndObjectTracking,Azim12ivs-datmo,Aldoma13icra-objectTracking,Peiliang18eccv-motionAndObjectTracking,Qiu19tro-vioObjectTracking}, we track a dense time-varying 
mesh model describing the shape of the human over time.  
Therefore, to create  a human node our \SPES needs to detect and estimate the shape of a human in the camera images, and then track the human over time. 
For shape estimation, we use the Graph-CNN approach of Kolotouros\setal~\cite{Kolotouros19cvpr-shapeRec}, which 
directly regresses  the 3D location of the vertices of an SMPL~\cite{Loper15tg-smpl} mesh model from a single image.
An example is given in Fig.~\ref{fig:humans}(a-b). More in detail, given a panoptic 2D segmentation, we crop the left camera image to 
 a bounding box around each detected human, and we use the approach~\cite{Kolotouros19cvpr-shapeRec} to get a 3D SMPL. 
We then extract the full pose in the original perspective camera frame (\cite{Kolotouros19cvpr-shapeRec} uses a  weak perspective camera model) 
using PnP~\cite{Zheng2013ICCV-revisitPnP}.   

To track a human, \ourThe \SPES builds a pose graph where each node is assigned the pose of the torso of the human at a discrete time. 
Consecutive poses are connected by a factor~\cite{Dellaert17fnt-factorGraph} modeling a zero velocity prior. 
Then, each detection at time $t$ is modeled as a prior factor on the pose at time $t$. 
For each node of the pose graph, \ourThe  \SPES also stores the 3D mesh
estimated by~\cite{Kolotouros19cvpr-shapeRec}. For this approach to work reliably, outlier rejection and data association become particularly important. The approach of \cite{Kolotouros19cvpr-shapeRec} often produces largely incorrect poses when the human is partially occluded.
 Moreover, in the presence of multiple humans, one has to associate each detection $d_t$ to one of the human pose 
 graphs $h\at{i}_{1:t-1}$ (including poses from time 1 to $t\!-\!1$ for each human $i=1,2,\ldots$). 
  To gain robustness, our \SPES (i) rejects detections when the bounding box of the human approaches the boundary of the image or is too small ($\leq 30$ pixels in our tests), and (ii) adds a measurement to the pose graph only when the human mesh detected at time $t$ is ``consistent'' with 
  the mesh of one of the humans at time $t\!-\!1$. 
  {To check consistency, we extract the skeleton at time $t-1$ (from the pose graph) and $t$ (from the current detection) and check that the 
  motion of each joint (Fig.~\ref{fig:humans}(c)) is physically plausible in that time interval (\ie we leverage the fact that the joint and torso motion cannot be arbitrarily fast). %
  We use a conservative bound of \maxJointDist on the 
  maximum allowable joint displacement in a time interval of \deltaT. If no pose graph meets the consistency criterion, we initialize a new pose graph with a single node corresponding to the current detection.}\veryOptional{\footnote{The SMPL parameters also describe the shape 
  of the body and could be potentially used as a descriptor for data association.}} 
  Besides using them for tracking, 
  we feed back the human detections to \KimeraSemantics, such that dynamic elements are not 
  reconstructed in the 3D mesh.
  We achieve this by only using the free-space information 
  when ray casting the depth for pixels labeled as humans, an approach we dubbed \emph{dynamic masking} (see results in Fig.~\ref{fig:dynamicMesh}).

\renewcommand{\myhspace}{\hspace{-6mm}}
\renewcommand{\mpw}{3.2cm}
\renewcommand{\myRate}{3.3cm}

\begin{figure}[h]
\vspace{-2mm}
	\begin{center}
	\begin{minipage}{\textwidth}
	\begin{tabular}{ccc}%
	\myhspace \hspace{3mm}
			\begin{minipage}{\mpw}%
			\centering%
			\includegraphics[height=\myRate]{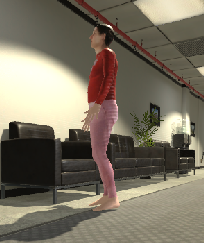} \\
			(a) Image
			\end{minipage}
		& \myhspace 
			\begin{minipage}{\mpw}%
			\centering%
			\includegraphics[trim=8mm 0mm 8mm 0mm, clip, height=\myRate]{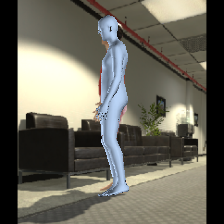} \\
			(b) Detection
			\end{minipage}
		& \myhspace 
			\begin{minipage}{\mpw}%
			\centering%
			\includegraphics[trim=120mm 40mm 20mm 110mm, clip,height=\myRate]{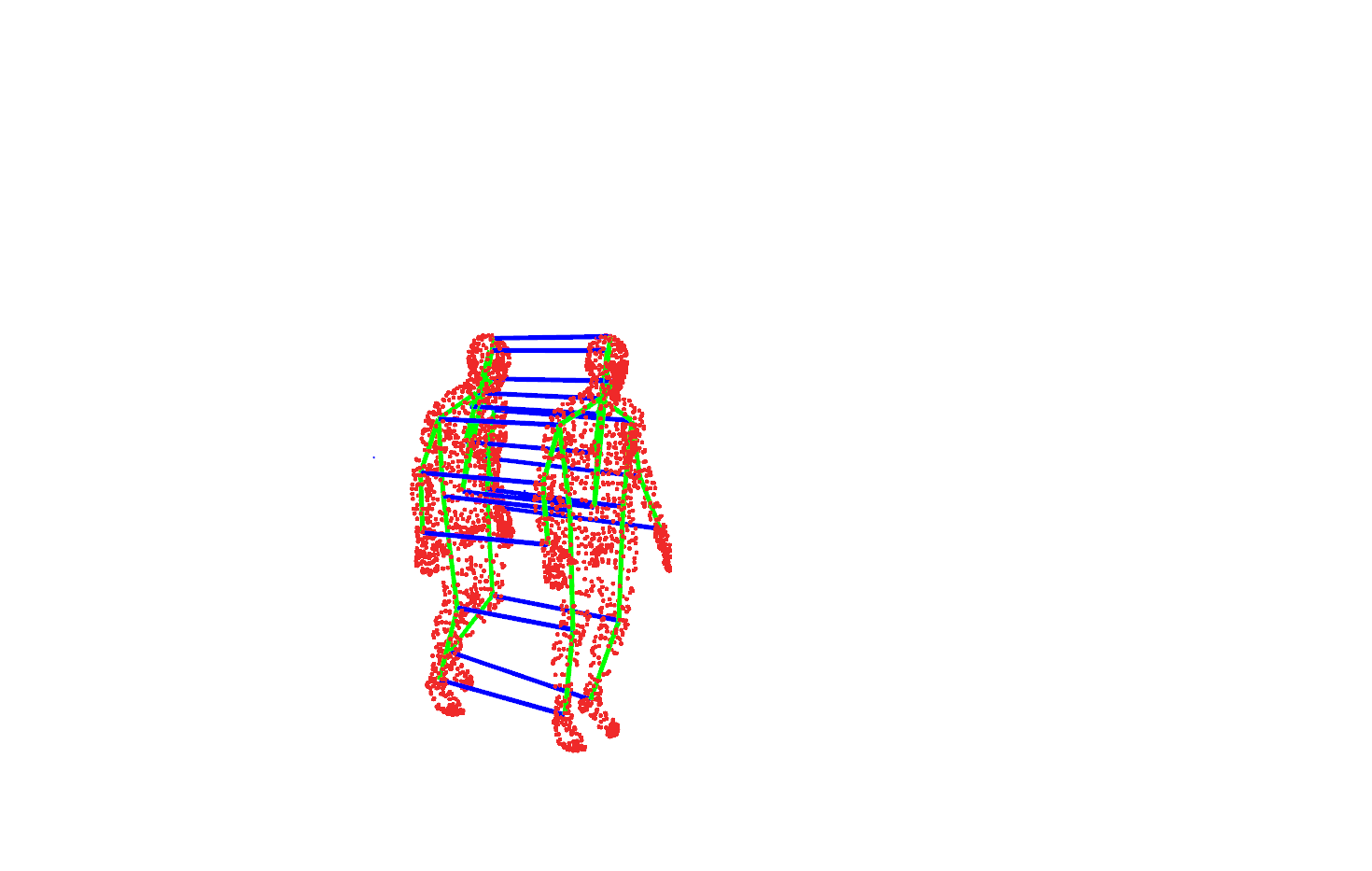} \\
			(c) Tracking
			\end{minipage}
		\end{tabular}
	\end{minipage}
	\begin{minipage}{\textwidth}
	\end{minipage}
	\vspace{-1mm} 
	\caption{{\bf Human nodes}: (a) Input camera image from Unity, (b) SMPL mesh detection and pose/shape estimation using~\cite{Kolotouros19cvpr-shapeRec}, (c) Temporal tracking and consistency checking on the maximum joint displacement between detections.
	 \label{fig:humans}}
	\vspace{-8mm} 
	\end{center}
\end{figure}

\subsection{From Mesh to Objects}
\label{sec:fromMeshToObjects}

Our spatial perception engine extracts static objects from the metric-semantic mesh produced by \Kimera.
 We give the user the flexibility to provide a catalog of CAD models 
 for some of the object classes. If a shape is available, 
 our \SPES will try to fit it to the mesh (paragraph ``Objects with Known Shape'' below), otherwise will only attempt to estimate 
 a centroid and bounding box (paragraph ``Objects with Unknown Shape'').

\myParagraph{Objects with Unknown Shape}
The metric semantic mesh from \Kimera already contains semantic labels.  
Therefore, \ourThe \SPES first exacts the portion of the mesh belonging to a given object class 
\toCheck{(\eg  chairs in~\FigFrontCover(d))}; this mesh potentially contains multiple object instances belonging to the same class.
Then, it performs Euclidean clustering using PCL~\cite{Rasu11icra} (with a distance threshold of {twice the voxel size used in \KimeraSemantics, which 
is $0.1$m)} to segment the object mesh into instances.
From the segmented clusters, \ourThe \SPES obtains a centroid of the object (from the vertices of the corresponding mesh), 
\toCheck{and assigns a canonical orientation with axes aligned 
with the world frame. Finally, it computes a bounding box with axes aligned with the canonical orientation.} 

\myParagraph{Objects with Known Shape}
For objects with known shape, \ourThe \SPES isolates the mesh corresponding to an object instance, {similarly to the unknown-shape case.} 
However, if a CAD model for that class of objects is given, 
\ourThe \SPES attempts fitting the known shape to the object mesh. 
This is done in three steps. First, 
we extract 3D keypoints from the CAD model of the object, and the corresponding object mesh from \Kimera.
The 3D keypoints are extracted by transforming each mesh to a point cloud (by picking the vertices of the mesh) and then 
extracting 3D Harris corners~\cite{Rasu11icra} with 
\toCheck{$0.15$m radius and $10^{-4}$ non-maximum suppression threshold.} 
Second, we match every keypoint on the CAD model with any keypoint on the \Kimera model. Clearly, this step produces many incorrect putative matches (outliers). Third, we apply a robust open-source registration technique, \TEASERpp~\cite{Yang20arxiv-teaser}, to find the best alignment between the point clouds in the presence of extreme outliers. 
The output of these three steps is a 3D pose of the object (from which it is also easy to extract an axis-aligned bounding box), 
see result in~\FigFrontCover(e).
\optional{Note that the object nodes implicitly enable to access the object shape at different \emph{levels of details}: 
the noisy and incomplete \Kimera mesh, the high-resolution CAD model, 
and the coarse bounding box.} %
\subsection{From Mesh to  Places, Structures, and Rooms}
\label{sec:fromMeshToStructures}

This section describes how our \SPES leverages existing techniques and implements simple-yet-effective methods to 
parse places, structures, and rooms from \Kimera's 3D mesh. %

\myParagraph{Places}
 \Kimera uses Voxblox~\cite{Oleynikova2017iros-voxblox} to extract a global mesh and an ESDF.
 We also obtain a topological graph from the ESDF using~\cite{Oleynikova18iros-topoMap}, where nodes %
sparsely sample  the free space, while edges represent straight-line traversability between two nodes.
We directly use this graph to extract the places and their topology (Fig.~\ref{fig:rooms}(a)).  
After creating the places, we associate each object and agent pose
to the nearest place to model a proximity relation.

\myParagraph{Structures}
\Kimera's semantic mesh already includes different labels for walls, ground floor, and ceiling, so isolating these three 
structural elements is straightforward (Fig.~\ref{fig:structure}). For each type of structure, we then compute a centroid, assign a canonical orientation (aligned with the world frame), and compute an axis-aligned bounding box.

\myParagraph{Rooms}
While floor plan computation is challenging in general, 
(i) the availability of a 3D \ESDF and 
(ii) the knowledge of the gravity direction given by \Kimera 
enable a simple-yet-effective approach to partition the environment into different rooms.
The key insight is that an horizontal 2D section of the 3D ESDF, cut below the level of the detected ceiling, is relatively unaffected by clutter in the room.
 This 2D section gives a clear signature of the room layout: the voxels in the section have a value of $0.3$m almost everywhere (corresponding to the distance to the ceiling), except close to the walls, where the distance decreases to $0$m.
  We refer to this 2D \ESDF (cut at $0.3$m below the ceiling) as an \emph{\ESDF section}.

 To compensate for noise, we further truncate the \ESDF section to distances above $0.2$m, such that small openings between rooms (possibly resulting from error accumulation) are removed. 
 The result of this partitioning operation is a set of disconnected 2D ESDFs corresponding to each room, that we refer to as \emph{2D \ESDF rooms}. 
Then, we label all the ``Places'' (nodes in Layer 3) that fall inside a 2D \ESDF room depending on their 2D (horizontal) position.
 At this point, some places might not be labeled (those close to walls or inside door openings).
 To label these, we use majority voting over the neighborhood of each node in the topological graph of ``Places'' in Layer 3; we repeat majority voting until all places have a label. 
 Finally, we add an edge between each place (Layer 3) and its corresponding room (Layer 4), see Fig.~\ref{fig:rooms}(b-c), and add an edge between two rooms (Layer 4) if there is an edge connecting two of its places (red edges in Fig.~\ref{fig:rooms}(b-c)). 
 We also refer the reader to 
 the video attachment.

\section{Experiments in Photo-Realistic Simulator}
\label{sec:experiments}

This section shows that the proposed \SPES 
(i) produces accurate metric-semantic meshes and robot nodes in crowded environments (Section~\ref{sec:exp-dynamicScenes}), 
(ii) correctly instantiates object and agent nodes (Section~\ref{sec:exp-humanAndObjects}), 
and (iii) reliably parses large indoor environments into rooms (Section~\ref{sec:exp-roomParsing}).

\myParagraph{Testing Setup} 
We use a photo-realistic Unity-based simulator to test our spatial perception engine in a 65m$\times$65m simulated office environment.
The simulator also provides the 2D panoptic semantic segmentation for \Kimera. 
Humans are simulated using
the realistic 3D models provided by the SMPL project~\cite{Loper15tg-smpl}. %
 \veryOptional{More specifically, (i) each agent chooses a random destination sampled on a circle of radius 25m 
 around its current position, (ii) the code computes the nearest collision-free location in the environment, 
and then (iii) each agent uses an off-the-shelf A$^*$ planner (available in Unity) 
to plan a path to the goal location. The process is repeated once the goal is attained.}
The simulator provides ground-truth poses of humans and objects, which are only used for benchmarking.
Using this setup, we create 3 large visual-inertial datasets, that we release as part of the \UnityHumans dataset~\cite{Rosinol20website-uHumans}.
The datasets, labeled as 
\toCheck{\UnityOne, \UnityTwo, \UnityThree, include 12, 24, and 60 humans, respectively.}
We use the human pose and shape estimator~\cite{Kolotouros19cvpr-shapeRec} out of the box, 
without  any domain adaptation or retraining. 

\subsection{Robustness of Mesh Reconstruction in Crowded Scenes}
\label{sec:exp-dynamicScenes}

Here we show that IMU-aware feature tracking and the use of a 2-point RANSAC in \Kimera enhance 
VIO robustness. Moreover, we show that this enhanced robustness, combined with \emph{dynamic masking} (Section~\ref{sec:fromVItoMeshAndAgents}), results in robust and accurate metric-semantic meshes in crowded environments.

\myParagraph{Enhanced VIO} 
Table~\ref{tab:robustVIO} reports the absolute trajectory errors of \Kimera with and without the 
use of 2-point RANSAC and when using 2-point RANSAC and IMU-aware feature tracking (label: \KimeraDVIO). 
Best results (lowest errors) are shown in bold.
The left part of the table (MH\_01--V2\_03) corresponds to tests on the (static) \Euroc dataset. 
The results confirm that in absence of {dynamic agents} the proposed approach performs on-par with the state of the art, 
while the use of 2-point RANSAC already boosts performance.
The last three columns  (\scenario{uH_01}--\scenario{uH_03}), however, \toCheck{show that in the presence of dynamic entities, 
the proposed approach dominates the baseline (\KimeraVIO).} %

\begin{table}[t!]
  \centering
  \caption{VIO errors in centimeters %
   on the \Euroc (MH and V) and \UnityHumans (uH) datasets.\vspace{-1mm}}
  \label{tab:robustVIO}
  \setlength{\tabcolsep}{3pt}
  \begin{tabular}{cccccccccccc|ccc}
  \toprule
  Seq. & \rotatebox[origin=c]{90}{MH\_01} &\rotatebox[origin=c]{90}{MH\_02} & \rotatebox[origin=c]{90}{MH\_03} & \rotatebox[origin=c]{90}{MH\_04}& \rotatebox[origin=c]{90}{MH\_05} & \rotatebox[origin=c]{90}{V1\_01} &  \rotatebox[origin=c]{90}{V1\_02} &  \rotatebox[origin=c]{90}{V1\_03}  &  \rotatebox[origin=c]{90}{V2\_01}  & \rotatebox[origin=c]{90}{V2\_02} & \rotatebox[origin=c]{90}{V2\_03} & \rotatebox[origin=c]{90}{\UnityOne} & \rotatebox[origin=c]{90}{\UnityTwo} & \rotatebox[origin=c]{90}{\UnityThree} \\ \midrule
  \specialcell[b]{5-point} & 9.3 & 10 & 11  & 42 & 21 & 6.7 & 12 & 17 & \textbf{5} & \textbf{8.1} & 30 & 92 & 145 & 160 \\
  \specialcell[b]{2-point} & 9.0 & {10} & \textbf{10} & {31} & \textbf{16} & {4.7} & \textbf{7.5} & \textbf{14} & 5.8 & 9 & \textbf{20} & 78 & {79} & {111} \\
  \specialcell[b]{ \textbf{\KimeraDVIOsuff}} & \textbf{8.1} & \textbf{9.8} & 14 & \textbf{23} & 20 & \textbf{4.3} & 7.8 & {17} & {6.2} & {11} & {30} & \textbf{59} & \textbf{78} & \textbf{88} \\ \bottomrule
  \end{tabular}
  \vspace{-3mm}
\end{table}

\renewcommand{\myhspace}{\hspace{0mm}}
\renewcommand{\mpw}{4.2cm}
\renewcommand{\myRate}{1}

\begin{figure}[h]
	\begin{center}
	\begin{minipage}{\textwidth}
	\begin{tabular}{cc}%
	\myhspace \hspace{-3mm}
			\begin{minipage}{\mpw}%
			\centering%
			\includegraphics[trim=400mm 6mm 500mm 50mm, clip,width=\myRate\columnwidth]{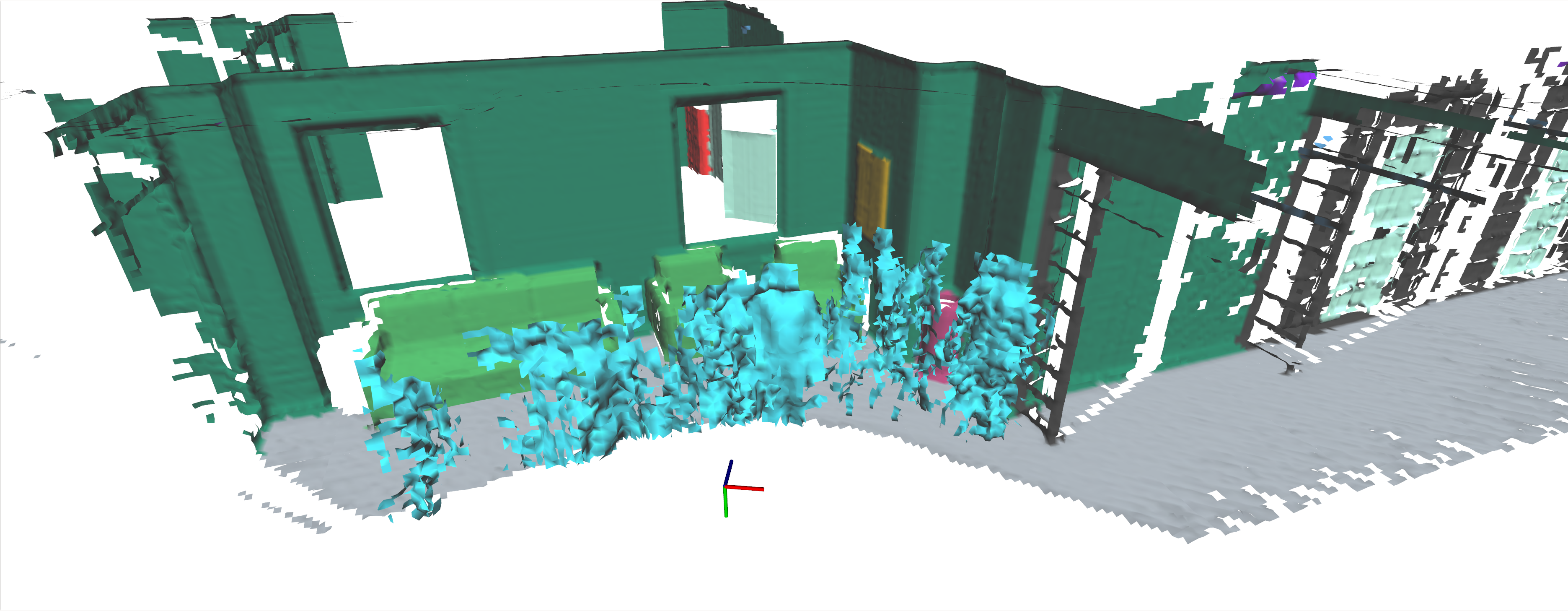} 
			\vspace{-18mm}\\
			\hspace{-1cm}(a) 
			\end{minipage}
		& \myhspace 
			\begin{minipage}{\mpw}%
			\centering%
			\includegraphics[trim=400mm 6mm 500mm 50mm, clip,width=\myRate\columnwidth]{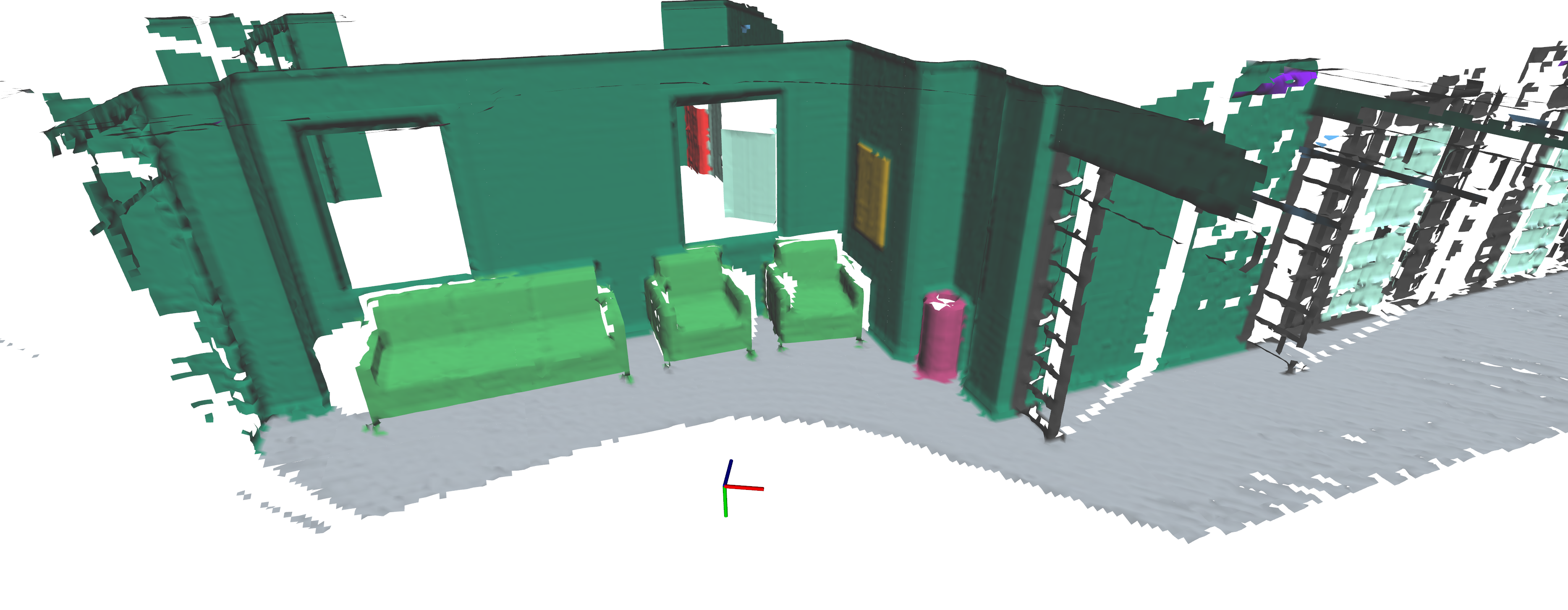} 
			\vspace{-18mm}\\
			\hspace{-1cm}(b) 
			\end{minipage}
		\end{tabular}
	\end{minipage}
	\begin{minipage}{\textwidth}
	\end{minipage}
	\vspace{-2mm} 
	\caption{3D mesh reconstruction (a) without and (b) with \emph{dynamic masking}.
	 \label{fig:dynamicMesh}}
	\vspace{-9mm} 
	\end{center}
\end{figure} 
\myParagraph{Dynamic Masking}
Fig.~\ref{fig:dynamicMesh} visualizes the effect of dynamic masking on \Kimera's metric-semantic mesh reconstruction.
Fig.~\ref{fig:dynamicMesh}(a) shows that without dynamic masking a human walking in front of the camera leaves 
a ``contrail'' (in cyan) and creates artifacts in the mesh. 
Fig.~\ref{fig:dynamicMesh}(b) shows that 
dynamic masking avoids this issue and leads to clean mesh reconstructions. 
Table~\ref{tab:mapAccuracy} reports the RMSE mesh error (see \emph{accuracy metric} in~\cite{Rosinol19icra-mesh}) 
with and without dynamic masking (label: ``with DM'' and ``w/o DM''). 
To assess the mesh accuracy independently from the VIO accuracy, we also report 
the mesh error when using ground-truth poses (label: ``GT Poses'' in the table), besides the results with the VIO poses
 (label: ``DVIO Poses'').
 The ``GT Poses'' columns in the table show that even with a perfect localization, the artifacts created by dynamic entities
 (and visualized in Fig.~\ref{fig:dynamicMesh}(a)) significantly hinder the mesh accuracy, while dynamic masking ensures highly accurate reconstructions. The advantage of dynamic masking is preserved when VIO poses are used.

\begin{table}[h!]
\vspace{-2mm}
  \centering
  \caption{Mesh error in meters with and without dynamic masking (DM).\vspace{-1mm} } %
  \label{tab:mapAccuracy}
  \begin{tabularx}{\columnwidth}{l *5{Y}}
    \toprule
    \multicolumn{1}{c}{Seq.} & \specialcell[b]{GT Pose \\ w/o DM} & \specialcell[b]{GT Poses \\ with DM} 
    & \specialcell[b]{\KimeraDVIOsuff~Poses \\ w/o DM}   & \specialcell[b]{\KimeraDVIOsuff~Poses \\ with  DM}    \\
    \bottomrule
    \UnityOne & 0.089 & 0.060 & 0.227 & 0.227 \\ 
    \UnityTwo & 0.133 & 0.061 & 0.347 & 0.301 \\
    \UnityThree & 0.192 & 0.061 & 0.351 & 0.335 \\
    \bottomrule
  \end{tabularx}
   \vspace{-4.5mm}
\end{table}

\subsection{Parsing Humans and Objects}
\label{sec:exp-humanAndObjects}

Here we evaluate the accuracy of human tracking and object localization on the \UnityHumans datasets.

\myParagraph{Human Nodes} 
Table~\ref{tab:humanAccuracy} shows the average localization error (mismatch between the torso estimated position and the ground truth) for each human on the \UnityHumans datasets. 
The first column reports the error of the detections produced by~\cite{Kolotouros19cvpr-shapeRec} (label: ``Single-img.''). 
The second column reports the error for the case in which we filter out detections when the human is only partially visible in the camera image, or when the bounding box of the human is too small ($\leq 30$ pixels, label: ``Single-img. filtered''). 
The third column reports errors with the proposed pose graph model discussed in~Section~\ref{sec:fromVItoMeshAndAgents} (label: ``Tracking''). 
 The approach~\cite{Kolotouros19cvpr-shapeRec} 
 tends to produce incorrect estimates when the 
  human is occluded.
Filtering out detections improves the localization performance, but %
occlusions due to objects in the scene still result in significant errors.
Instead, the proposed approach ensures accurate human tracking.

\begin{table}[h!]
\vspace{-2mm}
  \centering
  \caption{Human and object localization errors in meters.\vspace{-0mm}}
  \label{tab:humanAccuracy}
  \begin{tabularx}{\columnwidth}{l *6{Y}}
    \toprule
    & \multicolumn{3}{c}{\specialcell[b]{Humans}} & \multicolumn{2}{c}{\specialcell[b]{Objects}}
    \\
    \cmidrule(l{2pt}r{2pt}){2-4}
    \cmidrule(l{2pt}r{2pt}){5-6}
    \multicolumn{1}{c}{Seq.} & 
    \specialcell[b]{Single-img. \\ \cite{Kolotouros19cvpr-shapeRec}} 
    & \specialcell[b]{Single-img. \\ filtered} 
    & \specialcell[b]{Tracking \\ (proposed)}
    & \specialcell[b]{Unknown \\ Objects}  & \specialcell[b]{Known \\ Objects}   \\
    \midrule
    \UnityOne & 1.07 \optional{(2.07)} & 0.88 \optional{(1.36)} & 0.65 \optional{(0.69)} &  1.31 \optional{(2.21)}  & 0.20 \optional{(0.05)}\\ 
    \UnityTwo & 1.09 \optional{(1.79)} & 0.78 \optional{(0.94)} &0.61 \optional{(0.65)}  &  1.70 \optional{(2.22)}  & 0.35 \optional{(0.23)}\\
    \UnityThree & 1.20 \optional{(1.93)} & 0.97 \optional{(1.85)}&0.63 \optional{(0.73)}  &  1.51 \optional{(2.20)}  & 0.38 \optional{(0.20)}\\
    \bottomrule
  \end{tabularx}
  \vspace{-2mm}
\end{table}{} 
\myParagraph{Object Nodes} 
The last two columns of Table~\ref{tab:humanAccuracy} report the average localization errors for objects of unknown and known shape detected in the scene. %
In both cases, we compute the localization error as the distance between the estimated and the ground truth centroid of the object (for the objects with known shape, we use the centroid of the fitted CAD model).
We use CAD models for objects classified as ``couch''.
In both cases, we can correctly localize the objects,  while 
the availability of a CAD model further boosts accuracy.
\subsection{Parsing Places and Rooms}
\label{sec:exp-roomParsing}

The quality of the extracted places and rooms can be seen in Fig.~\ref{fig:rooms}. 
We also compute the average precision and recall for the classification of places into rooms.
The ground truth labels are obtained by manually segmenting the places.
For \scenario{uH_01} we obtain an average precision of $99.89\%$ and an average recall of $99.84\%$. 
 Incorrect classifications typically occur near doors, where room misclassification is inconsequential.

\section{Discussion: Queries and Opportunities} %
\label{sec:discussion}

We highlight the actionable nature of a 3D Dynamic Scene Graph 
by providing examples of queries it enables. 
\myParagraph{Obstacle Avoidance and Planning}  %
Agents, objects, and {rooms in our \DSG have a bounding box attribute.}
Moreover, the hierarchical nature of the \DSG ensures that bounding boxes at higher layers contain bounding boxes at lower layers 
(\eg the bounding box of a room contains the objects in that room). 
This forms a \emph{Bounding Volume Hierarchy} (BVH)~\cite{Larsson06cg-bvh}, which is extensively used for collision checking in computer graphics. BVHs provide readily available opportunities to speed up obstacle avoidance and motion planning queries
where collision checking is often used as a primitive~\cite{Karaman11ijrr-planning}.
\DSGs also provide a powerful tool for  high-level planning queries. %
For instance, the (connected) subgraph of places and objects in a \DSG can be used to 
issue the robot a high-level command (\eg object search~\cite{Joho11ras-objectSearch}), and the robot 
can directly infer the closest place in the \DSG it has to reach to complete the task, and can 
plan a feasible path to that place. 

The multiple levels of abstraction afforded by a \DSG 
have the potential to enable hierarchical and multi-resolution planning approaches~\cite{Schleich19rss-vinLevelsAbstraction,Larsson19arxiv-qSearch}, where a robot can plan at different levels of abstraction to save computational resources.
\myParagraph{Human-Robot Interaction} 
As already explored in~\cite{Armeni19iccv-3DsceneGraphs,Kim19tc-3DsceneGraphs}, a scene graph 
 can support user-oriented tasks, such as interactive visualization and \emph{Question Answering}. 
Our Dynamic Scene Graph extends the reach of~\cite{Armeni19iccv-3DsceneGraphs,Kim19tc-3DsceneGraphs} by 
(i) allowing visualization of human trajectories and dense poses (see visualization in the video attachment), 
and (ii) enabling more complex and time-aware queries such as 
``where was this person at time $t$?'', or 
``which object did this person pick in Room A?''. 
Furthermore, \DSGs provide a framework to model plausible interactions between agents and scenes \cite{Zhang19arxiv, Hassan19iccv, Pirk17tog, Monszpart19tog-imapper}.
We believe \DSGs also complement the work on natural language grounding~\cite{Kollar17arxiv-GGG},
 where one of the main concerns is to reason over the variability of
human instructions.

\myParagraph{Long-term Autonomy} 
\DSGs provide a natural way to ``forget'' or retain information in long-term autonomy. 
By construction, higher layers in the \DSG hierarchy are more compact and abstract representations of the 
environment, hence the robot can ``forget'' portions of the environment that are not frequently observed by simply pruning the 
corresponding branch of the \DSG. For instance, to forget a room in \FigFrontCover, we 
only need to prune the corresponding node and the connected nodes at lower layers (places, objects, etc.).
 More importantly, the robot can selectively decide which information to retain: 
 for instance, it can keep all the objects (which are typically fairly cheap to store), but can selectively forget the 
 mesh model, which can be more cumbersome to store in large environments.
 Finally, \DSGs inherit memory advantages afforded by standard scene graphs: if the robot detects $N$ instances of a known object (\eg a chair), 
 it can simply store a \emph{single} CAD model and cross-reference it in $N$ nodes of the scene graph; this simple observation 
 enables further data compression. %
\myParagraph{Prediction} 
The combination of a dense metric-semantic mesh model and a rich description of the agents allows  
performing short-term predictions of the scene dynamics and answering queries about possible future outcomes. 
For instance, one can feed the mesh model to a physics simulator and roll out potential high-level actions of the human agents;
 \optional{In general, we hope \DSG can help bridge the gap between robotics, computer vision, and graphics,  
 and enable the use of advanced graphics engines to support  decision-making tasks.}

\section{Conclusion}
\label{sec:conclusion}

We introduced \emph{3D Dynamic Scene Graphs} as a unified 
representation for actionable spatial perception, 
and presented the first \emph{\SPESlong} (\SPES) that builds a \DSG from sensor data in a fully automatic fashion. 
 We showcased our \SPES in a photo-realistic simulator, and discussed its {application to}
 several queries, including planning, 
human-robot interaction, data compression, and scene prediction.
This paper opens several research avenues. 
First of all, {many of the queries in Section~\ref{sec:discussion} involve nontrivial research questions and 
deserve further investigation.}
Second, more research is needed to expand the reach of \DSGs, for instance by developing algorithms that 
can infer other node attributes from data %
 (\eg material type and affordances for objects) or creating new node types for different environments (\eg outdoors). %
Third, this paper only scratches the surface in the design of spatial perception engines, %
 thus leaving many questions unanswered: is it advantageous to design \SPESs for other sensor combinations?
  Can we  estimate a scene graph incrementally and in real-time? 
  Can we design distributed \SPESs to estimate a \DSG from data collected by multiple robots?

\section*{Acknowledgments}

\thanks{This work was partially funded by ARL DCIST CRA W911NF-17-2-0181, ONR RAIDER N00014-18-1-2828, MIT Lincoln Laboratory, and
``la Caixa'' Foundation (ID 100010434), LCF/BQ/AA18/11680088 (A. Rosinol).} 

{\smaller %
 \bibliographystyle{abbrvnat}

}

\end{document}